\documentclass{article}

    \PassOptionsToPackage{numbers, compress}{natbib}

\usepackage[final]{neurips_2021}

\usepackage[utf8]{inputenc} 
\usepackage[T1]{fontenc}    
\usepackage{hyperref}       
\usepackage{url}            
\usepackage{booktabs}       
\usepackage{amsfonts}       
\usepackage{nicefrac}       
\usepackage{microtype}      
\usepackage{xcolor}         

\title{Building Legal Datasets}

\author{%
  Jerrold Soh Tsin Howe\\
  Centre for Computational Law, School of Law, Singapore Management University\\
  \texttt{jerroldsoh@smu.edu.sg} \\
}

\begin{document}

\maketitle

\begin{abstract}
Data-centric AI calls for better, not just bigger, datasets. As data protection laws with extra-territorial reach proliferate worldwide, ensuring datasets are \textit{legal} is an increasingly crucial yet overlooked component of ``better''. To help dataset builders become more willing and able to navigate this complex legal space, this paper reviews key legal obligations surrounding ML datasets, examines the practical impact of data laws on ML pipelines, and offers a framework for building legal datasets.
\end{abstract}

\section{Introduction}

Data-centric AI is about making better datasets. But what does ``better'' mean? Conventionally it has meant cheaper. That is, easier to crowdsource \cite{irani_turkopticon_2013}, generate \cite{forsyth_towards_2008}, augment \cite{dao_kernel_2019}, or broadly to collect \cite{paullada_data_2020}. \textit{Bigger} is also often better, as the rise of large language models suggest \cite{hendrycks_pretrained_2020, zhong_are_2021}. To statisticians, better typically means \textit{unbiased}, though ``bias'' is used differently from in the bias-variance tradeoff \cite{geman_neural_1992}, or in algorithmic bias \cite{friedman_bias_1996}. The growing ``responsible AI'' literature emphasizes that datasets are better when they are ethically and fairly sourced \cite{paullada_data_2020, hutchinson_towards_2021, rogers_just_2021}. This paper underscores \textit{legality} as one desideratum for ``better''. To this end, it reviews key legal obligations on data collection and use, examines the practical impact of data laws on ML pipelines, and offers a framework for thinking about data legality.

\section{When are datasets legal?}

Legal datasets may be understood broadly as datasets which are legally collected, retained, processed, \textit{and} disseminated. This fourfold categorization builds off Solove's classic taxonomy of privacy \cite{solove_taxonomy_2006}, and finds expression in a range of relatively new legislation worldwide. This notably includes the European Union's (EU's) \textit{General Data Protection Regulation} (GDPR) which came into force in 2018. Parallel to the GDPR are \textit{national} data laws, such as South Korea's \textit{Personal Information Protection Act} (passed in 2011), Singapore's \textit{Personal Data Protection Act} (passed in 2014) and, most recently, China's \textit{Personal Information Protection Law} (PIPL, August 2021). While the US does not presently have data legislation at the federal level, states like California, New York, and Massachusetts have passed data privacy acts. Further, legal scholars and courts have increasingly considered how pre-existing laws, such as copyright and anti-discrimination law, affect ML datasets \cite{sag_new_2019, mayson_bias_2019, gillis_big_2019}. As there are too many countries and variations to cover, I use the GDPR, PIPL, and California's \textit{Consumer Privacy Act} (CCPA, 2018) as case studies.

Although one jurisdiction's laws generally do not apply in another, modern data laws tend to have extra-territorial effect. Both the GDPR and PIPL apply as long as any personal data about persons in the EU/China is processed for any commercial or behavioral monitoring purposes (GDPR, Art 3; PIPL, Art 3). Likewise, Art 2 of the EU's proposed \textit{Artificial Intelligence Act} (AIA) expressly covers AI systems deployed in, or whose outputs are used, in the EU, regardless of where the providers and users of the system are. By contrast, the CCPA applies primarily to large businesses which ``do business in'' the state (CCPA, \S 1798.140). Thus, ML researchers and practitioners worldwide are now subject to foreign, and increasingly complex \cite{koops_trouble_2014}, data laws. Below I non-exhaustively review key legal obligations they impose. Note that ``legal'' here refers only to formal \textit{law}. This distinguishes my scope from (no less important) work on ``ethical'', ``fair'' or ``responsible'' AI \cite{paullada_data_2020, hutchinson_towards_2021, rogers_just_2021}. Despite clear overlaps, neither is a subset of the other. To illustrate, for some in certain states abortion is ethical yet illegal, for others elsewhere it is unethical yet legal.

\subsection{Collection}
\label{sec:when:consent}

Most centrally, data protection laws require informed consent before ``personal'' data may be obtained (GDPR, Arts 6--11; PIPL, Arts 13--17). The CCPA does not expressly require ``consent'', but businesses must inform consumers of the scope and purposes of data collected before collection (CCPA, \S 1798.100). The legality of numerous facial recognition datasets has been challenged for lack of consent \cite{noauthor_legality_2021, paullada_data_2020, obrien_facebook_2021}. Facial recognition clearly involves personal data because the task is to \textit{identify}. ``Personal data'' is, however, wider. Article 4 GDPR defines it as ``any information relating to an identified or identifiable natural person''. One is ``identifiable'' when they may be identified directly (i.e.\ by name) or indirectly. Names are not necessary; zip codes, gender, race, etc, could collectively identify. Indeed, Wong suggests that the EU's definition of personal data ``appears to be capable of encompassing all information in its ambit'', as EU courts have taken ``personal'' to include not only data \textit{about} a person, but also data which \textit{affects} them \cite{wong_delimiting_2019}.

The breadth of data laws explains why although most of ImageNet's \cite{deng_imagenet_2009} label classes do not target persons, its caretakers recently blurred out all human faces in the data, citing privacy concerns \cite{knight_researchers_2021}. A similar fate appears to have befallen the new Meta's facial recognition systems \cite{obrien_facebook_2021}. While most legal scrutiny has been on images, text, sound, and other modalities can also be ``personal''. One's forum posts, even if pseudonymous, could reveal much of their background. As such, dataset builders should be deliberate about obtaining consent even (or especially) when it is not obvious if the data is ``personal''.
 

\subsection{Retention}

A standard feature of legal consent is that consent may be withdrawn at any time. Data subjects may request to correct or erase their data (GDPR, Arts 16--17; PIPL, Arts 15, 16, 44--47; CCPA, \S 1798.105--106). Beyond consent, data controllers are also obliged to keep data in personally-identifiable form for no longer than necessary for its stated purposes (GDPR, Art 5(1)(e); CCPA, \S 1798.100(3)). Data that has served these purposes (say, the model has been trained) must be deleted or anonymized. However, given that data previously collected for one purpose can turn out useful for another, deletion may be quite undesirable for ML engineers. Anonymization is not much better, since preventing re-identification may require destroying most of a dataset's informative signals \cite{rocher_estimating_2019, xu_implications_2021}.

As such, prior thought should be given to delineating, and communicating, what the data will be used for. Conveying a specific purpose such as ``training ML models'' may not cover \textit{maintaining} or \textit{updating} the model post-deployment. Too general a purpose, such as ``for ML processing'' invites user suspicion and may fall outside the legal requirement that consent must be given in respect of ``specific'' purposes (GDPR, Art 6(1)(a); PIPL, Art 6).

\subsection{Processing}

Consent obligations surrounding data collection apply equally to data use. Further, data subjects have a right to be informed of and object to decisions ``based solely on automated processing'' (GDPR, Art 22; see also PIPL, Art 24). This legally advantages human-\textit{in-the-loop} systems. Beyond data protection laws, anti-discrimination laws in certain jurisdictions (e.g.\ US disparate treatment/impact laws; UK's \textit{Equality Act} 2010) may prohibit the use of protected attributes like race and gender for profiling \cite{hellman_measuring_2020}. This restricts the feature set which can legally be used for training ML models. Features highly co-linear with protected attributes may be indirectly prohibited as well.

While the obligations above may be more relevant to \textit{models} than to \textit{datasets}, laws can target the latter directly. Most prominently, the proposed Art 10 AIA stipulates that ``[t]raining, validation and testing data sets'' to be used in what the Act identifies as ``high-risk AI systems''  shall be ``relevant, representative, free of errors and complete''. This extends to having ``appropriate statistical properties'' regarding the system's target persons, and considering characteristics ``particular to the ... setting within which the high-risk AI system is intended to be used''. The draft AIA is in early stages and may take years and numerous amendments to come into force (if it does). Should it become law as is, it may effectively render data-centricity \textit{legally mandatory} for ``high-risk'' AI. Examples of high-risk AI enumerated in Annex III AIA non-exhaustively include systems for biometric identification, educational assessments, recruitment, credit scoring, law enforcement, and judicial decisions.

\subsection{Sharing and disclosure}

As data sharing or disclosure also constitutes processing, unauthorized disclosure is also a breach (GPDR, Arts 4(2); PIPL, Art 25). Thus, datasets with potentially personal information cannot be open-sourced without proper anonymization, even for research purposes. Another concern particular to large neural networks is the possibility that the network may memorize and leak personal information in the training data \cite{nasr_comprehensive_2019, chen_comprehensive_2020}. Personal information in datasets used for training large neural networks may therefore need to be removed \textit{before} training. 



\subsection{Research exemptions}
\label{sec:when:research}

The above obligations may be subject to limited research exemptions whose scope differs across jurisdictions \cite{mabel_research_2019}. For instance, both the GDPR and the CCPA regard subsequent scientific or statistical research as compatible with the initial purposes of the data collection for which consent was presumably obtained. This allows research to proceed without needing to ask for additional consent, subject to appropriate safeguards (GDPR, Arts 1(b) \& 89(1); CCPA, \S 1798.140(s)). This may have been sufficient to cover research applications of the ImageNet data (discussed above) without requiring anonymization. China's PIPL, however, does not have such an exemption.


\section{Implications on ML pipelines}

There is, in short, an expanding range of legal constraints on when and how data may be used. This has obvious implications for the ML community. Since \textit{legal} data is necessarily a subset of \textit{all} data, prioritizing legality seems to require sacrificing model performance. But less data is not always worse, especially if it also means less noise. More formally, if we think of ML broadly as seeking $argmax_{H} g(H, y)$, where the hypothesis $H(\theta; X, D)$ takes weights $\theta$ learned from features $X$ in dataset $D$, and $g(H, y)$ is a performance metric measured against (holdout) truth labels $y$, then legality constraints might be understood as follows:

$$argmax_{H} g(H, y) \quad \mathbf{s.t.} \quad X \subseteq X_{legal}; D \subseteq D_{legal}$$

with $X_{legal}$ and $D_{legal}$ respectively denoting the legally-permissible set of features and datasets. 

Formally framing the problem as such identifies three situations where legal constraints may not necessarily limit model performance. For brevity, we illustrate this with $D$, though similar logic applies with $X$. First, if $D_{legal}\!=\!D_\Omega$ (all data relevant to a task). That is, data laws have no practical effect on datasets in that area. For example, the task involves only cat detection and never implicates personal data. Second, if $D^*$, the theoretically optimal dataset, happens to be perfectly legal so that $D^*=D_{legal}$. Third, and least obviously, if the legally-constrained optimization problem produces \textit{better} performance than its unconstrained variant. This counter-intuitive result may occur in the real world because the law may force one to exclude noisy data (or features) that would otherwise have been included. In this sense, data laws, like other optimization constraints, may turn out to have a useful regularizing effect.


Moreover, in practice $g(H,y)$ is not the only metric to be optimized. Even assuming we care solely about economic value, profits, while correlated with (F1 score) performance, turn also on variables like user adoption and trust \cite{toreini_relationship_2019, gillath_attachment_2021, kerasidou_before_2021}. A perfectly accurate classifier that is never used generates no revenue. Fines also reduce profit. Ignoring legality provides another source of ``hidden debt'' in ML pipelines \cite{sculley_hidden_2015}. Thus. early investments in processes and practices for making legal datasets could yield better \textit{real world} performance, particularly in the long-run (where legal enforcement becomes more feasible). Apart from any obligation to follow the law just because it is law, there are practical reasons why the ML community should do so.




\section{Building legal datasets}



\begin{figure}
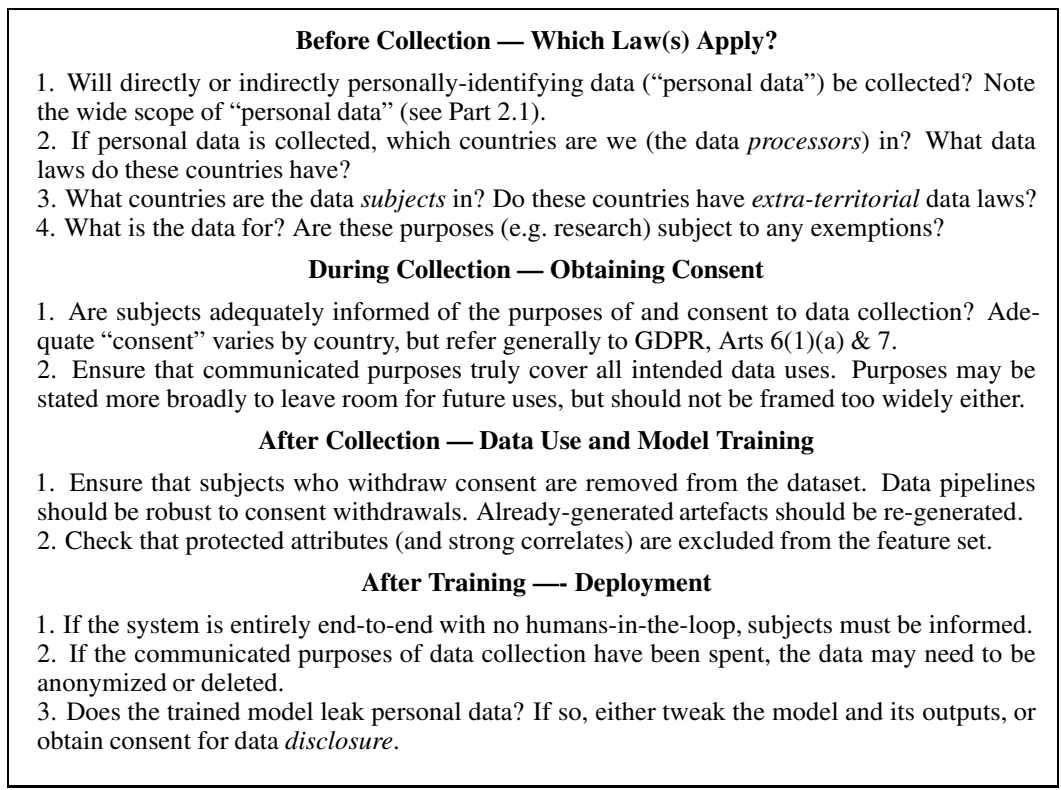

\framebox[\textwidth]{
\parbox{0.95\textwidth}{

\begin{center}
\textbf{Before Collection --- Which Law(s) Apply?}
\end{center}
1. Will directly or indirectly personally-identifying data (``personal data'') be collected? Note the wide scope of ``personal data'' (see Part \ref{sec:when:consent}).

2. If personal data is collected, which countries are we (the data \textit{processors}) in? What data laws do these countries have?

3. What countries are the data \textit{subjects} in? Do these countries have \textit{extra-territorial} data laws?

4. What is the data for? Are these purposes (e.g.\ research) subject to any exemptions?

\begin{center}
\textbf{During Collection --- Obtaining Consent}
\end{center}

1. Are subjects adequately informed of the purposes of and consent to data collection? Adequate ``consent'' varies by country, but refer generally to GDPR, Arts 6(1)(a) \& 7. 

2. Ensure that communicated purposes truly cover all intended data uses. Purposes may be stated more broadly to leave room for future uses, but should not be framed too widely either.

\begin{center}
\textbf{After Collection --- Data Use and Model Training }
\end{center}

1. Ensure that subjects who withdraw consent are removed from the dataset. Data pipelines should be robust to consent withdrawals. Already-generated artefacts should be re-generated.

2. Check that protected attributes (and strong correlates) are excluded from the feature set.

\begin{center}
\textbf{After Training ---- Deployment}
\end{center}
1. If the system is entirely end-to-end with no humans-in-the-loop, subjects must be informed.

2. If the communicated purposes of data collection have been spent, the data may need to be anonymized or deleted.

3. Does the trained model leak personal data? If so, either tweak the model and its outputs, or obtain consent for data \textit{disclosure}.
\\
}}
\caption{Suggested framework for dataset legality}
\label{fig:1}
\end{figure}

Complying with the intricate and growing web of data laws is non-trivial. The challenge is how we might turn motherhood calls for ``multi-disciplinary collaboration'' into actionable steps for ML researchers. The rise of ethics guidelines and responsible AI checklists \cite{zook_ten_2017, rogers_just_2021} offers one solution. In a sense, this involves, ethicists, sociologists, etc pre-computing complex, open-ended obligations into simpler, close-ended compliance heuristics for computer scientists. Following this trend, Figure \ref{fig:1} offers a framework for thinking about dataset legality. This builds on existing work that already incorporates some legal principles (e.g.\ \cite{, rogers_just_2021}) but differs in two ways. First, the framework focuses more on \textit{legality} and thus \textit{complements} responsible/ethical AI work. Second, while checklists and impact statements are generally backward-looking, encouraging researchers to justify choices already made, these considerations are forward-looking, encouraging researchers to think about legality at each stage of the ML process. This is crucial because legal errors, especially the need to obtain informed consent for processing, are expensive to rectify \textit{post facto} if not avoided \textit{ex ante}.

\section{Limitations}
\label{sec:limit}

All heuristics are wrong, but some are useful \cite{box_science_1976}. This paper does not cover all the legal obligations, duties, and exemptions affecting ML datasets. Nor can following the proposed framework completely guarantee legality (nor fairness or morality). Indeed, the corpus of legislation affecting ML datasets is set to grow, amid concerns that current laws offer insufficient safeguards \cite{wachter_data_2019, united_nations_high_commissioner_for_human_rights_right_2021}. The draft AIA, if and when passed, would significantly alter the AI and data governance landscape; other jurisdictions may follow suit with their own Acts. The minutiae of dataset legality should be fleshed out in future, lengthier work. The primary aim here is to spark discussion on when and why legal data is better data. Data-centric AI presents an opportunity for the ML community to build better datasets --- in all the technical, statistical, ethical, and legal senses of the word.

\begin{ack}
The authors disclose no funding sources nor competing interests.
\end{ack}



{
\small
\bibliographystyle{abbrvnat}
\bibliography{all}
}

\end{document}